\documentclass[letterpaper, 10 pt, conference]{ieeeconf}

\IEEEoverridecommandlockouts %

\usepackage{amsmath} %
\usepackage{amssymb} %
\usepackage{amsfonts}
\usepackage[sort,compress]{cite}
\usepackage[usenames,dvipsnames]{xcolor}
\usepackage{graphicx}
\usepackage{tabularx}
\usepackage{multirow}
\usepackage{mathtools}
\usepackage{esvect} %
\usepackage[]{siunitx}
\usepackage{caption}
\usepackage[pdftex, pdfstartview={FitV}, pdfpagelayout={TwoColumnLeft},bookmarksopen=true,plainpages = false, colorlinks=true, linkcolor=black, citecolor = black, urlcolor = black,filecolor=black , pagebackref=false,hypertexnames=false, plainpages=false, pdfpagelabels ]{hyperref}
\usepackage[T1]{fontenc} %
\usepackage{mathtools, cuted} %
\usepackage{scalerel}

\usepackage{balance} %
\usepackage{booktabs} %
\usepackage[export]{adjustbox} %

\newcolumntype{x}[1]{%
>{\raggedleft\hspace{0pt}}p{#1}}%

\usepackage{algorithm}
\usepackage[noend]{algpseudocode} %
\definecolor{commentclr}{RGB}{34, 139, 34}

\usepackage{tikz}

\usepackage{subcaption}
\DeclareCaptionLabelSeparator{periodspace}{.\quad}
\usepackage{makecell}

\newcommand{\ra}[1]{\renewcommand{\arraystretch}{#1}}

\usepackage{amsthm}

\theoremstyle{definition}

\usepackage{letltxmacro}
\LetLtxMacro\orgvdots\vdots
\LetLtxMacro\orgddots\ddots

\makeatletter
\DeclareRobustCommand\vdots{%
	\mathpalette\@vdots{}%
}
\newcommand*{\@vdots}[2]{%
	\sbox0{$#1\cdotp\cdotp\cdotp\m@th$}%
	\sbox2{$#1.\m@th$}%
	\vbox{%
		\dimen@=\wd0 %
		\advance\dimen@ -3\ht2 %
		\kern.5\dimen@
		\dimen@=\wd2 %
		\advance\dimen@ -\ht2 %
		\dimen2=\wd0 %
		\advance\dimen2 -\dimen@
		\vbox to \dimen2{%
			\offinterlineskip
			\copy2 \vfill\copy2 \vfill\copy2 %
		}%
	}%
}
\DeclareRobustCommand\ddots{%
	\mathinner{%
		\mathpalette\@ddots{}%
		\mkern\thinmuskip
	}%
}
\newcommand*{\@ddots}[2]{%
	\sbox0{$#1\cdotp\cdotp\cdotp\m@th$}%
	\sbox2{$#1.\m@th$}%
	\vbox{%
		\dimen@=\wd0 %
		\advance\dimen@ -3\ht2 %
		\kern.5\dimen@
		\dimen@=\wd2 %
		\advance\dimen@ -\ht2 %
		\dimen2=\wd0 %
		\advance\dimen2 -\dimen@
		\vbox to \dimen2{%
			\offinterlineskip
			\hbox{$#1\mathpunct{.}\m@th$}%
			\vfill
			\hbox{$#1\mathpunct{\kern\wd2}\mathpunct{.}\m@th$}%
			\vfill
			\hbox{$#1\mathpunct{\kern\wd2}\mathpunct{\kern\wd2}\mathpunct{.}\m@th$}%
		}%
	}%
}
\makeatother

\let\oldnl\nl%
\newcommand{\nonl}{\renewcommand{\nl}{\let\nl\oldnl}}%
\makeatother

\def\br{\mathbb R}

\def\sl{\mathcal{L}}

\newcommand\eqdef{\mathrel{\overset{\makebox[0pt]{\mbox{\normalfont\tiny def}}}{=}}}

\graphicspath{{figures/}}

\usepackage{svg}
\svgpath{{figures/}}
\title{\LARGE \bf
Global Localization in Unstructured Environments using \\% 
Semantic Object Maps Built from Various Viewpoints}

\author{Jacqueline Ankenbauer, Parker C.\ Lusk, Annika Thomas, Jonathan P.\ How%
	\thanks{J.\ Ankenbauer, P.\ C.\ Lusk, A.\ Thomas and J.\ P.\ How are with the Department of Aeronautics and Astronautics, Massachusetts Institute of Technology.
	    {\tt \{jpedlow, plusk, annikat, jhow\}@mit.edu}. }
    \thanks{Supported by The Boeing Company under Cooperative Agreement MRA\#2017-656, ARL DCIST under Cooperative Agreement W911NF-17-2-0181, and by UPenn under ONR Award 584551.}
}%

\begin{document}

\maketitle
\thispagestyle{plain}
\pagestyle{plain}

\begin{abstract}
We present a novel framework for global localization and guided relocalization of a vehicle in an unstructured environment.
Compared to existing methods, our pipeline does not rely on cues from urban fixtures (e.g., lane markings, buildings), nor does it make assumptions that require the vehicle to be navigating on a road network.
Instead, we achieve localization in both urban and non-urban environments by robustly associating and registering the vehicle's local semantic object map with a compact semantic reference map, potentially built from other viewpoints, time periods, and/or modalities.
Robustness to noise, outliers, and missing objects is achieved through our graph-based data association algorithm.
Further, the guided relocalization capability of our pipeline mitigates drift inherent in odometry-based localization after the initial global localization.
We evaluate our pipeline on two publicly-available, real-world datasets to demonstrate its effectiveness at global localization in both non-urban and urban environments.
The Katwijk Beach Planetary Rover dataset~\cite{hewitt2018katwijk} is used to show our pipeline's ability to perform accurate global localization in unstructured environments.
Demonstrations on the KITTI dataset~\cite{geiger2012kitti} achieve an average pose error of $3.8$\,m across all $35$ localization events on Sequence 00 when localizing in a reference map created from aerial images.
Compared to existing works, our pipeline is more general because it can perform global localization in unstructured environments using maps built from different viewpoints.
\end{abstract}

\section{Introduction}\label{sec:intro}

Global localization is the process of determining a vehicle’s pose (i.e., position and orientation) in its environment without an initial estimate.
While global navigation satellite system (GNSS) based methods have traditionally been used to provide global positioning in open settings with good satellite visibility, positioning quality quickly degrades in the presence of occlusion, multipath, or spoofing (e.g., in urban canyons, underground, or adversarial settings).
To solve this problem, methods have been proposed that leverage onboard measurements to localize the vehicle within a reference map.

\begin{figure}[t!]
	\centering
	\includegraphics[width=1.0\columnwidth]{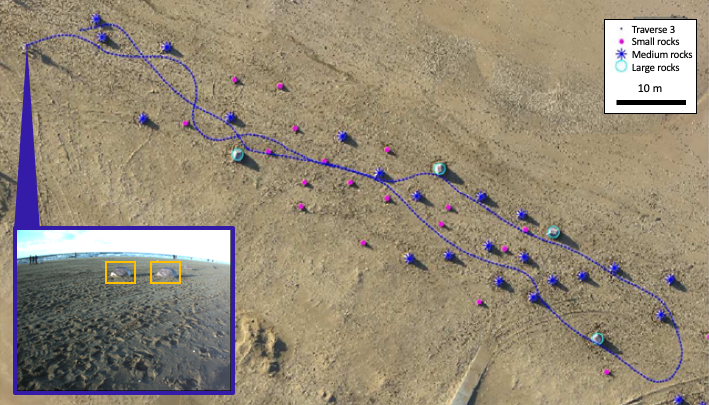}
	\caption{
	Global localization of a ground vehicle in an unstructured reference object map containing small, medium, and large rocks.
        The ground vehicle is localized by creating an object map of the rocks observed while driving (yellow bounding boxes in image) and registering them with the rocks in the reference map.
        With a reference map spanning roughly $1$\,km, $0.68$\,m accuracy has been achieved on the Katwijk dataset~\cite{hewitt2018katwijk}.
	}
	\label{fig:fig1}
	\vspace*{-0.3em}
\end{figure}

Current frameworks typically approach global localization with the assumption that the vehicle is on a road in an urban environment\cite{knights2022wild}.
Many methods localize a vehicle using OpenStreetMap (OSM) \cite{brubaker2015map, floros2013openstreetslam, ye2022crossview,cho2022openstreetmap, vysotska2017improving,yan2019global}, which is freely-available and memory efficient, but is limited to urban settings and mapping a new area requires significant effort.
Methods using semantics often focus primarily on structures existing only in urban environments (e.g., lane markings, traffic signs, and road structure \cite{pink2008visual, javanmardi2017towards,wang2022ltsr,brubaker2015map}) and begin with a non-uniform prior, i.e., assuming the vehicle can only be on a road\cite{miller2021any,wang2022ltsr}.
In practice however, applications such as military reconnaissance missions or search and rescue missions cannot make these assumptions.
Global localization in an unstructured environment requires localizing with minimal data because information such as road structure and lane markings is not present, for example in the environment shown in Fig.~\ref{fig:fig1}.
The search space is also increased because the vehicle could be in any pose within the reference map.

Changes in the environment such as lighting differences, seasonal changes, or objects being created, moved, or removed cause discrepancies between reference and vehicle maps, which is challenging for current methods.
High-density maps are high-quality and can lead to very low position errors, but they are expensive to generate and prone to becoming outdated \cite{gawel20173D}.
Furthermore, map size becomes increasingly important when considering sharing information between vehicles or a central database. %

Reference maps created from a different viewpoint than the vehicle (e.g., localizing a ground vehicle in an aerial map or using maps built from opposing views) is also an important capability and may be required based on map availability.
However, image-based methods which use descriptors such as visual bag of words~\cite{galvez2012bags} suffer from viewpoint changes.
These methods may fail when localizing a ground vehicle driving north in a reference map built from a ground vehicle driving south in the same area~\cite{lowry2015visual}.

To address these issues, we present a framework capable of globally localizing a vehicle in an unstructured environment using a reference map created from an arbitrary viewpoint.
Correspondences between objects in the local vehicle map and in the reference map are made by exploiting the geometric consistency of potential associations~\cite{lusk2021clipper}.
Informed by object semantics, we reduce the space of potential object associations and efficiently identify the largest set of geometrically consistent associations using a maximum clique solver~\cite{rossi2013parallel}.
The vehicle's pose is then found by registering the objects in the local vehicle map with their associated reference map objects.
Importantly, leveraging geometric consistency in such a graph-based manner enables our pipeline to achieve localization in unstructured maps from various viewpoints and with robustness against outliers in both the vehicle and reference maps.
Thus, as long as objects can be identified and reconstructed in both the vehicle and reference maps, global localization can be achieved.
While globally localizing in an unstructured environment is a challenging problem due to decreased amount of available data, an increased search space, and potentially-outdated maps, our pipeline has been specifically designed to succeed in these environments.

In summary, the contributions of this work are:
\begin{itemize}
    \item A global localization framework robust to outliers and viewpoint changes due to its graph-based object association formulation and use of compact semantic maps. %
    \item A framework capable of localizing in unstructured environments. 
 With no prior assumptions on the existence of an urban setting, the same pipeline has been demonstrated to successfully localize in unstructured environments such as the Katwijk Beach Planetary Rover dataset~\cite{hewitt2018katwijk}.
    \item A guided relocalization mode to continually correct the pose estimate after global localization in order to reduce effects of drift.
    \item A demonstration of successful localization and guided relocalization achieving state-of-the-art performance on real data from the KITTI dataset using a reference map from aerial images captured on a different date with many outliers.
\end{itemize}

\section{Related Work}

Global localization has close ties to literature on loop closure detection, place recognition, and image retrieval.
We review the most recent and related work in these domains.

\subsection{Image-Based Methods}
Appearance-based methods use images for localization by finding the most visually similar image in the reference database to the locally captured image~\cite{lowry2015visual}.
Visual similarity is typically assessed based on low-level information such as color and reflectance values~\cite{veronese2015re,vora2020aerial} or visual features/descriptors\cite{senlet2011framework,noda2010vehicle}.
Early works such as \cite{chen2011city,li2010location} use local feature descriptors to compare and match images taken from different perspectives.
Majdik et al.~\cite{majdik2013mav} use such features with simulated images from Google Street View to match against images from a quadrotor flying through an urban environment.
Methods based on low-level features are impacted the most by changes in the environment such as illumination or seasonal changes and, more importantly, many fail under extreme viewpoint difference between the vehicle and reference images (e.g., aerial-ground).

\subsection{Cross-View Methods}
Cross-view methods, which localize a ground vehicle in aerial/satellite imagery, have been specifically developed to handle extreme viewpoint differences.
Current state-of-the-art methods~\cite{shi2019spatial,rodrigues2021these,wang2021each} are learning-based and use a Siamese network architecture~\cite{kim2017satellite,tian2020cross} to return a coarse localization (accuracy of hundreds of meters) across a very large area (e.g., city-wide).
When coupled with particle filters, these techniques can provide a higher accuracy (tens of meters) in geo-tracking applications~\cite{downes2022city}.
Cross-view algorithms are typically designed for the extreme air-ground viewpoint difference, but may not be directly applicable for other viewpoint variations (e.g., ground-ground viewed from opposite directions).
Air-ground localization can also be achieved by other techniques, such as~\cite{wolcott2014visual,gawel20173D,barsan2020learning}, which can obtain highly accurate localization (centimeter-level) by exploiting a high-definition point cloud map of the environment.
However, while accurate, these methods do not work with image maps and require dense point clouds.

\subsection{Semantic-Aided Methods}
Semantic-aided methods leverage semantic information to assist with localization.
Some methods compactly identify objects, their location, and potentially other characteristics to create reference and vehicle maps to ultimately localize within each other~\cite{stenborg2018long,liu2019global,wang2022ltsr}.
Other methods use image segmentation~\cite{miller2021any,kim2019fusing}, semantic lidar point cloud matching~\cite{song2022semantic,segmatch2017}, general vertical structures~\cite{kummerle2011large,wang2017flag,wang2022ltsr}, lane markings~\cite{pink2008visual,javanmardi2017towards}, or buildings~\cite{matei2013image,senlet2014hierarchical,tian2017cross}.
Semantic maps are often summarized with descriptors such as random walk descriptors~\cite{lin2021topology,liu2019global}, histograms~\cite{zhu2020gosmatch}, or structural appearance~\cite{kim2021scan}, allowing the vehicle’s local observations to be compared with previously seen objects in the reference map.
Other semantic-aided methods use OpenStreetMap, which is readily available and requires little memory.
These methods compare observed roads~\cite{brubaker2015map, floros2013openstreetslam, ye2022crossview}, buildings \cite{cho2022openstreetmap, vysotska2017improving}, or both \cite{yan2019global} to determine where the vehicle is inside the reference map.
Overall, most semantic-aided methods are restricted to working only in urban or suburban settings.

\subsection{Unstructured Environments}
Most works within global localization and loop closure are not suited for non-urban settings due their rigid reliance on urban semantic information (e.g., buildings, roads, lane markings) or their need for rich features within images (e.g., appearance-based methods).
There have been works which address the difficulty in successfully running a SLAM system due to the lack of features and roughness in the road~\cite{yang2020multi, ji2018cpfg}.
Specific to localization, works have used topological maps~\cite{ort2018autonomous}, wheel odometry combined with visual orientation tracking~\cite{grimes2009efficient}, or lidar point clouds~\cite{ren2021lidar} in order to refine a GPS estimate.
Global localization in GPS-denied environments has been achieved by methods using lidar~\cite{gawel20173D}, stationary anchors within the region~\cite{stoll2017gps}, and binary ground-nonground distinction~\cite{viswanathan2016vision,viswanathan2014vision}.
Despite success with global localization, these methods are either restricted by the size of dense point clouds, have requirements for external hardware in the field (e.g., anchors), are not robust to structural changes, or assume the vehicle is on an off-road trail.

\subsection{Placement of This Work}
This work uses semantic object maps and geometric consistency in order to be view-invariant and robust to structural changes in the environment.
The generality of the classes being used and assumptions being made (i.e., no reliance on roads) allow this framework to successfully operate in unstructured environments.
Furthermore, this method addresses the issue of relocalization after global localization to mitigate effects of drift.
Few works specifically demonstrate drift reduction during egomotion tracking~\cite{floros2013openstreetslam,yan2019global}.

\section{Global Localization Pipeline} \label{sec:framework}

\subsection{System Overview}
Our pipeline has two operating modes: global localization and guided relocalization.
The pipeline starts in the global localization mode, wherein the vehicle searches for its global pose within a provided reference map.
We emphasize that in this mode, no prior information is leveraged (e.g., no initial guess and no assumption that a vehicle is restricted to roads).
Once a candidate transformation between the vehicle's local observations and reference map is accepted (see Section~\ref{subsec:global_loc}), global localization is achieved and the pipeline switches to guided relocalization.
The guided relocalization mode continually updates the accepted transformation by leveraging past information in order to reduce the drift of the SLAM system (see Section~\ref{subsec:guided_reloc}).

To identify candidate transformations between the vehicle and reference maps, the pipeline finds corresponding objects between the two maps.
The vehicle map $\mathcal{M}_\mathrm{veh}$ and reference map $\mathcal{M}_\mathrm{ref}$ consist of objects $o_i =(u_i, c_i)$, represented by their 3D centroid $u_i \in \br^3$ and a class, $c_i\in \mathcal{C}$, where $\mathcal{C}$ is a set of classes (e.g. parking spaces, boulders, etc.) known a priori.
The pipeline constructs $\mathcal{M}_\mathrm{veh}$ online by detecting objects at each timestep and reconstructing their centroids using onboard sensors.
After the objects at timestep $t_n$ are classified and reconstructed, the objects are compared to all objects seen in previous timesteps $t=t_0,\dots,t_{n-1}$ and all objects of the same class within a specified radius are fused together.
Drift in the trajectory estimate contributes to object reconstruction inaccuracies, so registration is performed on only the $r$ most recently seen objects, denoted $\mathcal{M}_\mathrm{veh}^r \subseteq \mathcal{M}_\mathrm{veh}$.

The reference map can either be constructed offline or be updating in real-time.
As the pipeline is searching over the entire reference map during global localization, we split the reference map into $k$ submaps with a specified level of overlap to increase computational efficiency.
All of the resulting submaps, $\mathcal{M}^i_{\mathrm{ref}} \subset \mathcal{M}_{\mathrm{ref}}$, $i = 1, \dots, k$, are input into the registration module.
In contrast, after global localization is achieved, the reference map can be strategically constrained to objects in close proximity to the vehicle map. %
This restricted reference map $\mathcal{M}_\mathrm{ref}^\mathrm{res}$ is input into the registration module during guided relocalization.

\subsection{Registration}\label{subsec:registration}
Robust registration is a core component of the proposed framework.
We use a graph-based formulation to solve the registration problem by finding the largest set of geometrically consistent objects that match between the reference submaps and vehicle map.
Denoting the association that matches the points $p_i$ and $q_i$ by $a_i =(p_i, q_i)$, two associations $a_i$ and $a_j$ are considered \textit{geometrically consistent} if and only if the distance between the points is preserved, i.e., $\| p_i - p_j \| = \| q_i - q_j \|$.
In practice, due to noise and inaccuracies, a threshold $\epsilon$ is set to consider associations consistent when $d(a_i, a_j) \eqdef | \, \| p_i - p_j \| - \| q_i - q_j \| \,| < \epsilon$. 
Now, by denoting the set of associations between objects of the same class in the reference and vehicle maps as
${A \eqdef \{ (o_i, o_j ): \, (u_i,c_i) \in \mathcal{M}_{\mathrm{ref}},\, (u_j,c_j) \in \mathcal{M}^r_{\mathrm{veh}},\, c_i = c_j \}}$, the problem of finding the largest set of consistent associations, $A^*_{\text{c}}$, can be defined formally as 
\begin{gather} \label{eq:maxclique}
	\begin{array}{ll}
		\underset{A_{\text{c}} \subset A}{\text{maximize}} & | A_{\text{c}} |
		\\
		\text{subject\ to} & d(a_i, a_j) < \epsilon, ~ \forall_{a_i,a_j \in A_{\text{c}}}.
	\end{array}
\end{gather}
 
\begin{figure}[t]
	\centering
	\includegraphics[width=0.99\columnwidth]{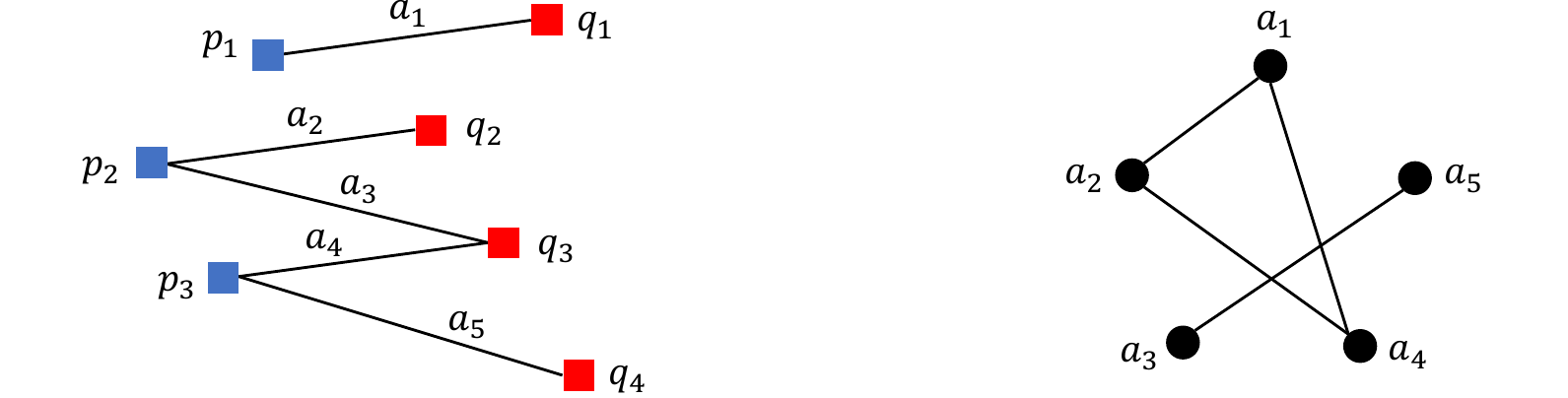}
	\caption{Maximum clique formulation for registration. (Left) Points $p$ and $q$ are matched by associations $a$. (Right) 
	Graph with nodes representing associations and edges indicating their consistency (i.e., associations with (nearly) identical distances between their endpoints).
	The largest clique ${A^*_{\text{c}} = \{a_1,a_2,a_4\}}$ is the largest set of consistent associations.}
	\label{fig:max_clique}
\end{figure}

Problem \eqref{eq:maxclique} can be modeled as a graph whose vertices represent associations and edges represent consistent associations.
The optimal solution is equivalent to the maximum clique of the graph, as illustrated in Fig.~\ref{fig:max_clique}.
Although typically NP-hard, finding the maximum clique can be solved relatively quickly for sparse graphs (resulting from many inconsistent associations created by an all-to-all scheme) using the parallel maximum clique (PMC) algorithm~\cite{rossi2013parallel}.

The registration module periodically registers $\mathcal{M}^r_{\mathrm{veh}}$ to each of the reference map submaps, i.e., $\mathcal{M}^i_\mathrm{ref}$ 
for global localization or $\mathcal{M}^\mathrm{res}_\mathrm{ref}$ for guided relocalization.
During global localization, we use an all-to-all association scheme within objects of the same class, i.e., we initially associate an object in each reference submap to every object in the vehicle map of the same class.
During guided relocalization, we leverage previously identified associations from the last relocalization event by restricting these objects to be associated only with each other.
The remaining objects are associated using an all-to-all association scheme to allow additional associations to be identified.

Solving~\eqref{eq:maxclique} provides the maximum set of consistent associations despite many outlier associations generated by the all-to-all association scheme.
These associations are then used in the least-square fitting of matched objects via Arun's method~\cite{arun1987least}, which gives the optimal transformation ${T_\mathrm{cand} \in \mathrm{SE}(3)}$ that registers $\mathcal{M}_\mathrm{veh}^r$ to $\mathcal{M}_\mathrm{ref}$.

\begin{figure}[t!]
	\centering
	\begin{subfigure}[b]{0.99\columnwidth}
		\centering
            \includeinkscape[pretex=\scriptsize,width=\columnwidth]{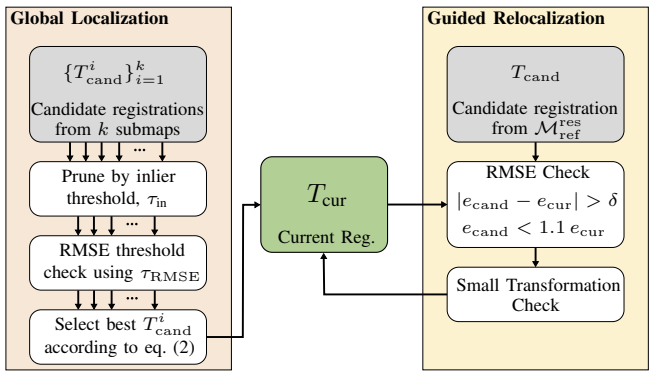}
	\end{subfigure}
	\vspace*{-1.0em}
	\caption{Flow chart for accepting or rejecting a new registration in the global localization and guided relocalization modes.}
	\vspace*{-0.2em}
	\label{fig:flow}
\end{figure}

\subsection{Global Localization}\label{subsec:global_loc}
The data and decision making flow of our pipeline is detailed in Fig.~\ref{fig:flow}.
During global localization, candidate registrations $\{T^i_\mathrm{cand}\}_{i=1}^k$ are identified for each of the $k$ submaps $\{\mathcal{M}^i_\mathrm{ref}\}_{i=1}^k$.
For a given candidate registration $T^i_\mathrm{cand}$, the number of inlier associations identified by problem~\eqref{eq:maxclique} is denoted as $a_i$.
Because registrations with few associations are less likely to be reliable (e.g., due to perceptual symmetries or the anticipation of a changed environment), candidates with less than $\tau_\mathrm{in}$ inlier associations are rejected.
We use $\mathbb{V}$ to denote the candidates which pass the inlier association threshold.

The quality of the $i$-th registration is evaluated using the root-mean-square error (RMSE), denoted $e_i$, which measures the differences between objects in the transformed full vehicle map and their nearest neighbors in the reference map.
This value provides insight into how well the vehicle map as a whole aligns with the reference map, as opposed to only considering how well the objects associated using $A_c^*$ are matched.
An RMSE threshold $\tau_\mathrm{RMSE}$ is used to check that there is at least one candidate transform of sufficient quality.
Importantly, $\tau_\mathrm{RMSE}$ increases as the distance traveled increases to account for the distortion in the map due to drift in the trajectory estimate.
If no candidate registration meets both the $\tau_\mathrm{in}$ and $\tau_\mathrm{RMSE}$ thresholds, the pipeline waits for new candidate transformations and repeats the process.
If, however, at least one candidate registration passes these thresholds, the best registration is selected by
\begin{gather} \label{eq:global_loc}
	\begin{array}{ll}
		i^* = & \underset{i\in\mathbb{V}}{\text{argmax}} \; a_i \\
		& \text{subject to} \quad e_i \leq (1+\alpha) \,\underbar{$e$}, %
	\end{array}
\end{gather}
where $\underbar{$e$}=\min_{j\in1,\dots,k}\{e_j : a_j \geq \tau_\mathrm{in}\} \leq \tau_\mathrm{RMSE}$ is the best RMSE value in the set of valid transformations and $0 < \alpha \ll 1$.
Thus, the result of \eqref{eq:global_loc} is the candidate transformation with the most associations, provided that the RMSE value is close to the best RMSE value.
In simpler words, if multiple transformations have similar RMSE near or below the threshold, the number of inliers is the best indicator of which transformation is accurate.
Once a registration is accepted, it is stored as the current transformation $T_\mathrm{cur}\gets T^{i^*}_\mathrm{cand}$.
Then, the pipeline switches to the guided relocalization mode where $T_\mathrm{cur}$ will be frequently updated.

\subsection{Guided Relocalization}\label{subsec:guided_reloc}
Guided relocalization is used to frequently update $T_\mathrm{cur}$, the current transformation between the local and global coordinate frames.
This is needed because as the vehicle moves, the trajectory estimation process accumulates drift.
The criteria for accepting the candidate transformation $T_\mathrm{cand}$ is detailed in the right of Fig.~\ref{fig:flow}.

During guided relocalization, a candidate registration is compared to the current registration $T_\mathrm{cur}$ to determine if the candidate registration will be accepted.
Unlike the global localization criteria, the new candidate registration does not have a required number of inlier associations because the framework is already confident in the approximate transformation between coordinate frames.
The RMSE value is calculated for both the current accepted registration ($e_\mathrm{cur}$) and the candidate registration ($e_\mathrm{cand}$).
These values are calculated as defined in Section~\ref{subsec:global_loc}, but using the $r' \geq r$ most recently seen objects in the vehicle map such that $\mathcal{M}_\mathrm{veh}^{r}\subseteq\mathcal{M}_\mathrm{veh}^{r'}$
Calculating the RMSE values with more vehicle map objects than were used to find the candidate transformation provides a better assessment of the quality of each transformation.
To accept $T_\mathrm{cand}$, the two RMSE values must be sufficiently different ($| e_{\mathrm{cand}} - e_{\mathrm{cur}} | > \delta$) and the candidate registration's value must be similar or smaller than the current registration's value ($e_{\mathrm{cand}} \leq (1+\alpha) \,e_{\mathrm{cur}}$).

The final criteria to accept the candidate transformation is that $T_\text{cand}$ must be similar enough to $T_\text{cur}$ in both translation and orientation.
The transformation similarity requirements loosen to account for drift as the distance traveled since the last accepted registration increases.
If accepted, $T_\text{cur}~\gets~T_\text{cand}$ and the process begins again when the next candidate registration is provided by the registration module.

\section{Experimental Evaluations} \label{sec:expr}
We evaluate our framework by localizing a ground vehicle in reference maps for both the Katwijk Beach Planetary Rover dataset~\cite{hewitt2018katwijk} and the KITTI dataset~\cite{geiger2012kitti}.
Our pipeline is implemented in C++ using ROS~\cite{quigley2009ros} and runs in real-time on an Intel i9 CPU with 64 GB RAM and a NVIDIA RTX 3080 GPU for object detection.
The Katwijk dataset is used to demonstrate the pipeline’s ability to localize a vehicle in an unstructured environment and its view-invariant property.
The experiment objectives for the KITTI dataset are to showcase the pipeline's ability to handle reference maps created from various viewpoints, robustness to outliers, and accuracy compared to other methods.

\begin{figure}[t!]
	\centering
	\includegraphics[trim={1mm 1mm 2mm 0},clip,width=0.99\columnwidth]{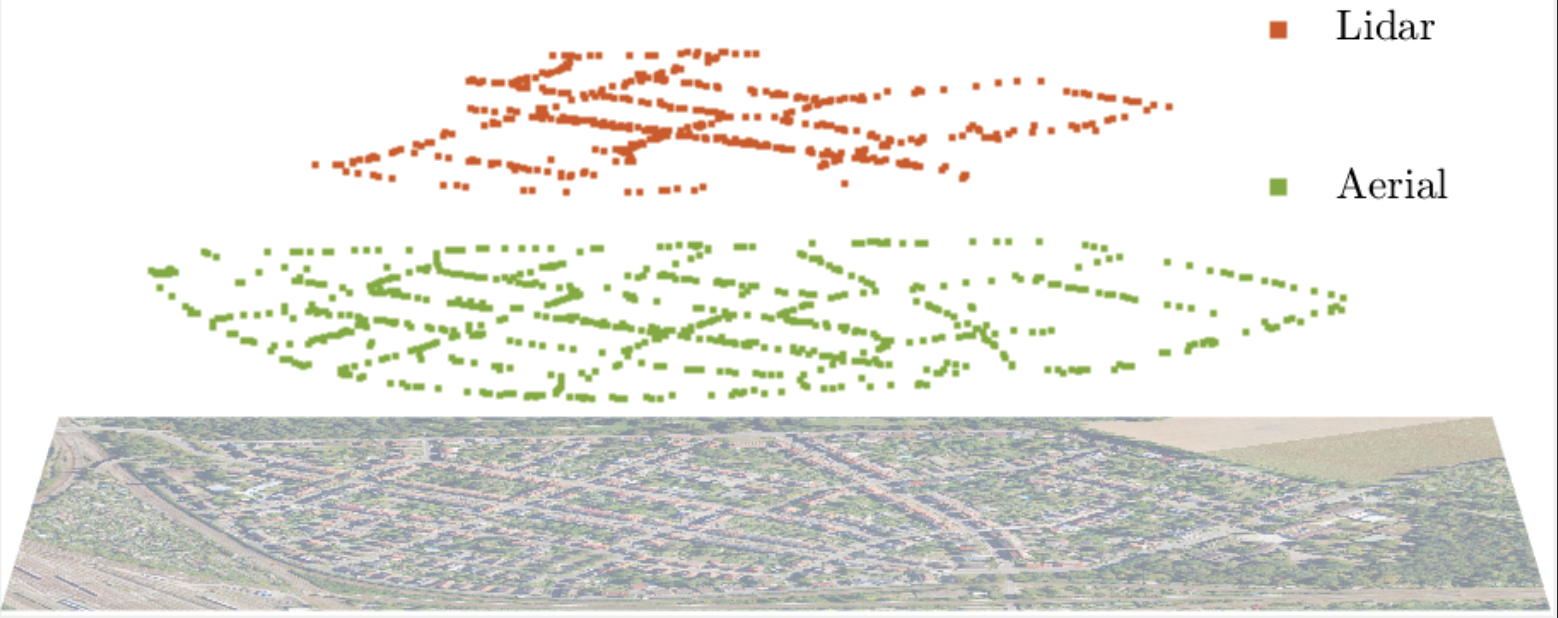}
	\vspace*{-0.5em}
	\caption{Reference object maps corresponding to KITTI Sequence 00 constructed using lidar scans (ground view), and Google Satellite georeferenced images (aerial view). Each square represents a semantic object such as a parking space or traffic sign. The bottom image is for reference.}
	\label{fig:maps}
	\vspace*{-0.5em}
\end{figure}

\subsection{Katwijk Dataset Experimental Setup}

The Katwijk Beach dataset provides a challenging scenario to test our global localization pipeline.
A rover was used to collect data while driving on a beach where small, medium, and large artificial rocks were placed in arbitrary locations along its route (see Fig.~\ref{fig:fig1}).
Using the stereo camera data (\emph{LocCam}) captured onboard the rover, rocks are detected and classified by size.
Rock detection is performed by reconstructing a 3D point cloud, removing the ground plane, and keeping point clusters that meet certain size criteria (the size of potential rocks is known a priori).
The 3D stereo point clouds and bounding boxes drawn around each rock were created before running the pipeline and saved in a ROS bag file.
The messages were then published in real-time as the pipeline runs.
In real-time the point cloud is projected onto each image and the points that lie within each bounding box are said to correspond to that object.
The distance to each object is calculated to be the median distance to the 3D points corresponding to that object.
The median distance is used because it is assumed some points may lie in the background of the bounding box, not on the object itself.
The final 3D estimate of the object centroid is taken to be the center of the bounding box projected into 3D space by the estimated distance to the object.

Due to the challenging nature of the Katwijk dataset, off-the-shelf visual odometry packages (e.g., ORB-SLAM3~\cite{ORBSLAM3_TRO}) failed to estimate the robot's trajectory.
In addition, the dataset does not provide a ground truth trajectory, so we generated coarse ground truth poses for each camera frame by interpolating the high-precision RTK GPS measurements.
For this reason, our generated ground truth was used in our pipeline and our experiments focus on testing the global localization capability.

Algorithm parameters are tuned on Traverse 1, Part 1 and the same values are used for other sequences.
In particular, we use a threshold of $\epsilon = 1.5$\,m  for registration (defined in Section~\ref{subsec:registration}) and we set the RMSE threshold value to $\tau_\mathrm{RMSE}~=~2$\,m.
We do not restrict the size of the vehicle object map (i.e., $\mathcal{M}^r_\mathrm{veh}=\mathcal{M}_\mathrm{veh}$) given how few rocks the vehicle sees.
For all segments of Traverses 1 and 2, a minimum of  $\tau_\mathrm{in}=8$ inliers are required.
However, this parameter is loosened to $\tau_\mathrm{in}=6$  for Traverse 3, as we can achieve high confidence of an accurate registration with less inliers because of the significantly lower number of objects in the reference map ($45$ as opposed to $212$).

\subsection{Unstructured Environments}
Using rocks classified by size, we demonstrate that our pipeline is able to globally localize the rover in an unstructured environment.
Traverses 1, 2, and 3 of the Katwijk dataset are split into $8$, $6$, and $5$ five-minute segments, respectively.
We tested on each of these segments and reported the five best results in Table~\ref{tbl:katwijk}.
Many segments do not achieve global localization because the rover does not see a sufficient number of rocks.
Two segments fail due to misclassifications and harsh lighting.
Even with these challenges, the registration error for global localization was as low as $0.58$\,m on Traverse 3, Part 4, when localizing in a reference map spanning approximately $110$\,m and as low as $0.68$\,m on Traverse 1, Part 1, on a reference map spanning roughly $1$\,km.  Furthermore, in Traverse 1, Parts 1 and 8, the rover only needed to identify $8$ and $9$ objects in order to localize in a map of $212$ objects.

\begin{table}[t] %
\scriptsize
\centering
\caption{
    Error statistics describing the pipeline's performance in unstructured environments on the Katwijk dataset.  Errors are reported in 2D.
}
\vspace*{-0.3em}
\setlength{\tabcolsep}{3.5pt}
\begin{tabular}{c c c c c c c c c c c c c c c c c c c c }
	\toprule
	\makecell{Traverse} & \makecell{Part} && \makecell{ Position \\ Error $[$m$]$ } && \makecell{Objects to\\ Localize $[$\#$]$} && \makecell{Objects in\\ Ref Map $[$\#$]$} && \makecell{Length of Ref \\ Map $[$m$]$} \\ 
	\toprule
	\makecell{$1$ \\$1$\\$3$ \\ $3$ \\ $3$} &  \makecell{$1$\\$8$\\$1$\\$2$\\$4$} && \makecell{$0.68$\\$0.97$\\$0.65$\\$1.4$\\$0.58$} &&  \makecell{$8$\\$9$\\$9$\\$12$\\$7$} && \makecell{$212$\\$212$\\$45$\\$45$\\$45$} && \makecell{~$1000$\\~$1000$\\~$110$\\~$110$\\~$110$}\\
	\bottomrule
\end{tabular}
\vspace*{-0.3em}
\label{tbl:katwijk}
\end{table}

\subsection{Viewpoint Variations}
In addition to localizing in the ground truth object map, Katwijk Traverse 3 was used to create a reference map from the extreme opposite viewpoint as the vehicle’s view.
Given that Traverse 3 is an ``out-and-back'' trajectory (see Fig.~\ref{fig:fig1}), the first half of the traverse (parts 1, 2, and 3) is used to create a reference map into which the second half is localized.
In other words, the two halves of the traverse see the same objects, but from an extreme difference in viewpoint.
This scenario is challenging to image-based methods, which are likely to fail due to sensitivity to viewpoint~\cite{lowry2015visual}.

In our pipeline, the object map representation and maximum clique based association formulation cause the framework to be view-invariant and we are able to localize with $1.2$\,m accuracy within a reference map spanning approximately $110$\,m.  The error comes from inaccuracies in the trajectory and object centroid reconstruction.

\subsection{KITTI Dataset Experimental Setup}
Experiments on the KITTI dataset enable us to compare to other methods and to demonstrate our pipeline's robustness to outliers.
To build the vehicle's object map, we use the stereo implementation of ORB-SLAM3~\cite{ORBSLAM3_TRO} for odometry estimation and YOLO~\cite{redmon2018yolov3,bjelonicYolo2018} for object detection.
For each sequence, two reference maps are considered.
One is built by taking the median values of points from the SemanticKITTI~\cite{behley2019semantickitti} point cloud, a semantically-labeled lidar scan captured from a ground viewpoint.
The other is created by manually annotating Google Satellite images using QGIS~\cite{QGIS_software} (the annotation can be automated by classifiers trained for aerial/satellite images~\cite{Ding_2019_CVPR, li2022oriented}).
The aerial and lidar reference maps for Sequence 00 can be seen in Fig.~\ref{fig:maps}.
The only object classes used are parking spaces and traffic signs, although most of objects are parking spaces.
Our classifier identifies cars as a proxy for parking spaces, but this leads to noisy estimates because not every parking space is occupied by a car and not every car is located in a parking space.
Furthermore, since the parking spots occupied by cars change over time, using semantic object maps from different dates further stresses our algorithm's robustness to outliers.
Sequences 00, 02, 06, 07, and 09 of the KITTI dataset are tested due to the number of semantic objects and the lack of symmetry.
For each of these sequences, SemanticKITTI~\cite{behley2019semantickitti} identifies greater than $100$ stationary cars and traffic signs and the objects in the reference map do not contain high levels of symmetry, as symmetry in surrounding areas leads to failure in global localization (see~\ref{subsec:discussion}).

Algorithm parameters are tuned on Sequence 00.
In particular, we use a threshold of $\epsilon = 2.5$\,m  for registration and require at least $\tau_\mathrm{in}=12$ inlier associations to accept a registration while restricting the size of the vehicle object map to the last $r=75$ seen objects.
We begin the RMSE threshold value at $\tau_\mathrm{RMSE}=6$\,m, though this threshold increases by $2$\,m for every $500$\,m traveled to account for vehicle map distortion due to drift in the trajectory estimate.
Additionally, the RMSE value is calculated using only parking space objects because the sparsity of traffic signs leads to large RMSE values.
No submaps are used for the lidar reference maps, but the aerial reference maps are split into either $k=2$ or $k=4$ submaps with no overlap, depending on the total number of reference map objects.
Reference maps split into two submaps (Sequences 02, 06, and 09) were split in half along the y-axis.

\subsection{Accuracy of Localization Events}
There are two primary sources of error in our pipeline: noisy object maps and noisy registrations.
Inaccuracies in the object maps are caused by errors during 3D centroid reconstruction and local pose estimates.
The 3D centroid reconstructions are either hand-labeled and prone to human error (aerial reference map) or estimated using point clouds generated by either lidar sensors (lidar reference map) or ORB-SLAM3 tracked features (vehicle map).
The accuracy of vehicle pose estimates depends on various factors such as sensors, number of loop closures, and drift from the SLAM module.
Incorrect pose estimates distort the object map and directly influence the localization accuracy in our pipeline.
In order to increase accuracy of the pipeline, localization events (i.e., receiving pose corrections) must be frequent and accurate.
After global localization, infrequent relocalizations would allow drift to accumulate between events and contribute toward inaccurate pose estimates across the entire sequence.

\begin{figure}[t!]
\centering
\begin{subfigure}[b]{0.99\columnwidth}
    \centering
    \includegraphics[trim = 0mm 0mm 0mm 0mm, width=1\textwidth]{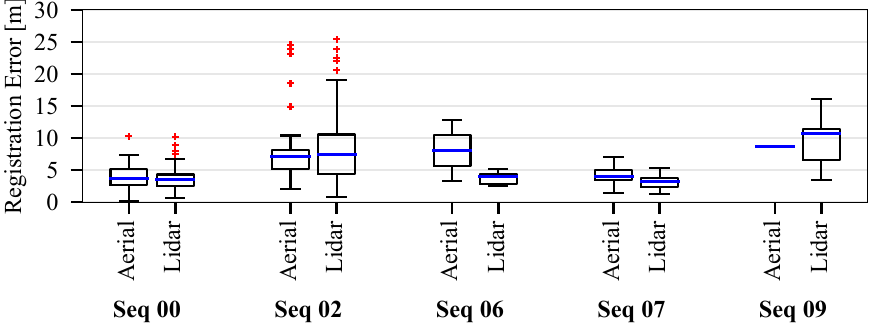}
\end{subfigure}
\caption{
    Error statistics for registration accuracy of each KITTI sequence and each reference map (aerial and lidar).
    Registration accuracy is defined as the pose estimate error at each localization event. 
    Errors are comparable across reference map cases, illustrating our framework's robustness to outliers and viewpoints.
    Errors are reported in 2D for the aerial case and 3D for the lidar case to mirror the dimension of the reference maps.
}
\vspace*{-0.3em}
\label{fig:boxplot}
\end{figure}

\begin{figure}[t!]
\centering
\begin{subfigure}[b]{0.99\columnwidth}
    \centering
    \vstretch{1}{\includegraphics[trim = 8mm 3mm 13mm 0mm, clip, width=1\textwidth]{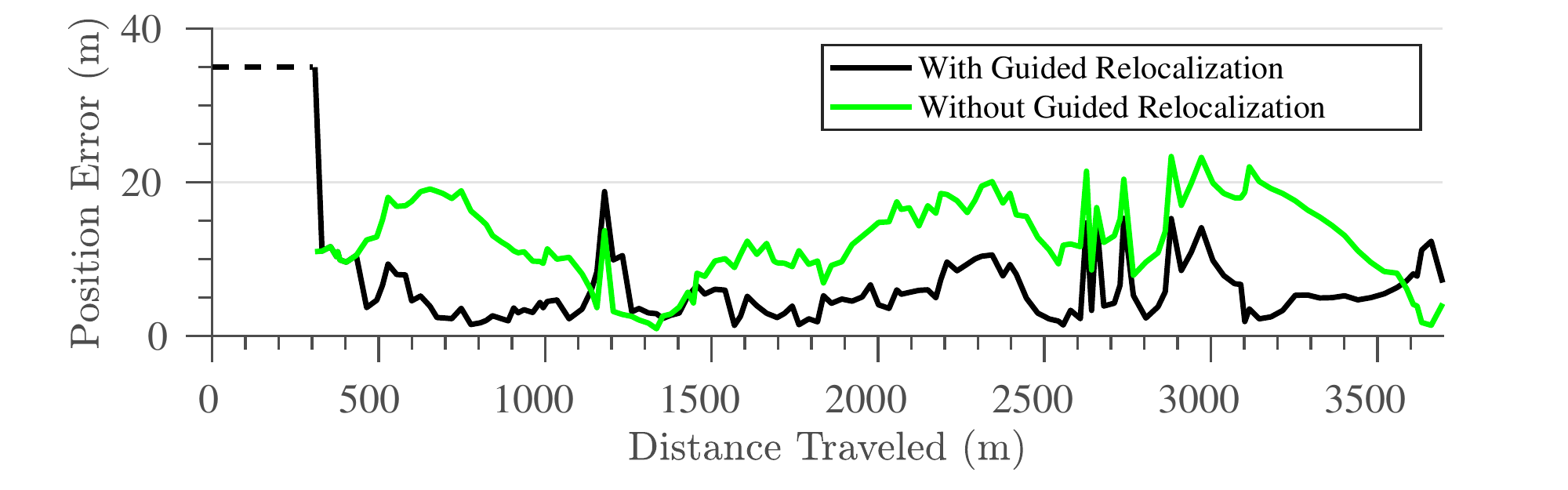}}
\end{subfigure}
\vspace*{-1.0em}
\caption{
    Estimated pose error with and without guided relocalization when localizing the KITTI Sequence 00 ground vehicle in an aerial reference map.
    The black line demonstrates the full ability of the pipeline whereas the green line is calculated as if the only accepted transformation is the initial global localization transformation.
}
\vspace*{-0.2em}
\label{fig:ablation}
\end{figure}

To quantify the quality of accepted registrations, we plot the pose estimate error of relocalization events in Fig.~\ref{fig:boxplot}.
These error statistics demonstrate accuracy of each localization event in each of the tested sequences.
It can be observed that Sequences 00 and 07 have the lowest registration error with averages of $3.8$\,m and $4.1$\,m when localizing in the aerial reference map.
These errors are smaller than the average position error for \cite{yan2019global} and \cite{floros2013openstreetslam} by a factor of 10 (see Table~\ref{tbl:comparison}), though methods such as \cite{miller2021any} and \cite{brubaker2015map} have superior accuracy on the KITTI dataset.
In addition to highly accurate average registration error, tight bounds on the error for Sequences 00, 06, and 07 demonstrate that the registrations are consistently accurate as a whole.
Sequence 02, however, does contain many outliers, which is attributed to a large amount of drift in the trajectory.

\begin{table}[!b]
\scriptsize
\centering
\caption{
Error statistics of KITTI ground vehicle localization in aerial and lidar reference maps using only parking spaces and traffic signs as semantic objects.
The reported position and orientation errors are in 2D for the aerial case and 3D for the lidar case to mirror the dimension of the reference maps.
}
\setlength{\tabcolsep}{1.25pt}
\ra{1.1}
\sisetup{detect-weight=true,detect-inline-weight=math}
\begin{tabular}{
    @{}l@{}c
    S[table-format=2.1]
    S[table-format=1.1]
    S[table-format=4]
    S[table-format=4.0]
    S[table-format=3.0]
    S[table-format=2.0]
    S[table-format=3.0]
    @{}
}
\toprule
&  \makecell{KITTI \\ Seq. \\ $[$\#$]$}
& {\makecell{Average \\ Position \\ Error $[$m$]$}}
& {\makecell{Average \\ Orientation \\ Error $[$deg$]$}}
& {\makecell{Distance to\\  Localize\\ $[$m$]$}}
& {\makecell{Trajectory\\ Length\\ $[$m$]$}}
& {\makecell{Objects to\\ Localize \\ $[$\#$]$}}
& {\makecell{Object \\Outliers \\ $[$\%$]$}}
& {\makecell{Objects in \\Ref. Map \\ $[$\#$]$}} \\ 
\toprule
\multirow{5}{*}{\rotatebox[origin=c]{90}{aerial ref. map}}
& 00 & 6.0 & 1.6 & 257 & 3724 & 71 & 80 & 942 \\
& 02 & 10.7 & 0.7 & 883 & 5067 & 87 & 82 & 471 \\
& 06 & 7.9 & 0.6 & 1120 & 1233 & 88 & 66 & 741 \\
& 07 & 4.3 & 2.9 & 454 & 695 & 105 & 81 & 942 \\
& 09 & 10.1 & 1.2 & 1362 & 1705 & 147 & 80 & 493 \\
\midrule
\multirow{5}{*}{\rotatebox[origin=c]{90}{lidar ref. map}}
& 00 & 5.1 & 2.3 & 209 & 3724 & 58 & 52 & 543 \\
& 02 & 10.2 & 1.4 & 763 & 5067 & 73 & 42 & 315 \\
& 06 & 6.8 & 1.2 & 551 & 1233 & 45 & 55 & 119 \\
& 07 & 3.6 & 1.6 & 153 & 695 & 50 & 51 & 180 \\
& 09 & 11.4 & 1.1 & 956 & 1705 & 94 & 64 & 167 \\
\bottomrule
\end{tabular}
\label{tbl:kitti_performance}
\end{table}

\begin{table*}[!t] %
\scriptsize
\centering
\caption{
	Aerial-ground localization comparison on the KITTI benchmark for position error and localization time. Dashed line ``$-$'' indicates not reported or not localized successfully.
}
\vspace*{-0.3em}
\setlength{\tabcolsep}{2.7pt}
\begin{tabular}{l c c c c c c c c c c c c c c c c c c }
	\toprule
	\multirow{3}{*}{Approach} & 
	\multirow{3}{*}{\makecell{Vehicle\\ Map}} && 
	\multirow{3}{*}{\makecell{Reference\\ Map}} &&
	\multicolumn{2}{c}{Seq 00} &&
	\multicolumn{2}{c}{Seq 02} &&
        \multicolumn{2}{c}{Seq 06} &&
        \multicolumn{2}{c}{Seq 07} &&
	\multicolumn{2}{c}{Seq 09} \\ 
	\cmidrule{6-7}\cmidrule{9-10} \cmidrule{12-13} \cmidrule{15-16} \cmidrule{18-19}
	 &  &&  
	 && \tiny{\makecell{Error $[$m$]$}} & \tiny{\makecell{Time $[$sec$]$}}  
	 && \tiny{\makecell{Error $[$m$]$}} & \tiny{\makecell{Time $[$sec$]$}} 
      && \tiny{\makecell{Error $[$m$]$}} & \tiny{\makecell{Time $[$sec$]$}} 
      && \tiny{\makecell{Error $[$m$]$}} & \tiny{\makecell{Time $[$sec$]$}} 
	 && \tiny{\makecell{Error $[$m$]$}} & \tiny{\makecell{Time $[$sec$]$}} \\
	\toprule
	Miller  \cite{miller2021any}  & L+C && OP  &&  $2.0$ & $54.6$ && $9.1$ & $71.5$ && $-$ & $-$ && $-$ & $-$ && $7.2$ & $75$ \\
	Yan \cite{yan2019global} & L && OSM &&  $>10$ & $-$ && $-$&$-$ && $>10$ & $-$ && $>10$ & $-$ && $>10$ & $-$ \\
	Brubaker \cite{brubaker2015map} & S && OSM && $2.1$ &$22$ && $4.1$ & $26$ && $-$ & $-$ && $1.8$ & $26$ && $4.2$ & $24$ \\
        Floros \cite{floros2013openstreetslam} & C && OSM && $>10$ & $-$ && $>20$ & $-$ && $-$ & $-$ && $-$ & $-$ && $-$ & $-$  \\
	\textbf{Ours} & GT && OP &&  $3.5$ & $39$  &&  $4.4$ & $72$   &&  $1.5$ & $93$ &&  $2.3$ & $60$ &&  $3.0$ & $75$  \\
	\textbf{Ours} & S  && OP &&  $6.0$ & $36$  &&  $10.7$ & $78$  &&  $7.9$ & $102$ &&  $4.3$ & $66$ &&  $10.1$ & $135$\\
	\bottomrule
\end{tabular}
\label{tbl:comparison}
\\ [0.2em]
C: Monocular Camera, ~ S: Stereo Camera, ~ L: Lidar, ~ GT: Ground Truth,  
OP: Orthophoto, ~ OSM: OpenStreetMap
\vspace*{-0.5em}
\end{table*}

The importance of guided relocalization is underscored when comparing the error at each localization event to the error at each timestep in the sequence.
For Sequence 00, the mean localization event errors are $3.8$\,m and $3.7$\,m across $35$ and $65$ localization events for the aerial and lidar reference maps, respectively.
In contrast, when considering the average error across the entire trajectory after global localization, the mean errors for Sequence 00 are $6.0$\,m and $5.1$\,m.
The larger error is attributed to accumulated drift between relocalization events.
To visualize the importance of frequent guided relocalization, Fig.~\ref{fig:ablation} demonstrates the estimated pose error with and without guided relocalization on KITTI Sequence~00.
High error around $1200$\,m and $2600$\,m are due to poor ORB-SLAM pose estimates during turns.
Overall, with global localization only, the average error is $12.5$\,m, but with guided relocalization, the average error is halved to $6.0$\,m.

Despite these challenges, the overall performance (see Table~\ref{tbl:kitti_performance}) and localization event accuracy (see Fig.~\ref{fig:boxplot}) for each of the reference maps are comparable.
In comparing the boxplots, it is important to recognize that the number of relocalization events differs between localizing in the aerial and lidar reference maps for the same sequence.
Localizing in the aerial reference map is more challenging and thus, takes a longer distance to localize for each sequence, which typically leads to less localization events (e.g., localizing Sequence 09 in the aerial reference map results in only one localization event).
Sequences 00, 02, and 07 demonstrate the most localization events, and the median errors are most similar across the two reference maps.
The similar statistics illustrate that the framework is invariant to viewpoints and changes in the environment.

\subsection{Discussion} \label{subsec:discussion}
Table~\ref{tbl:comparison} lists our evaluation results of localizing the KITTI ground vehicle in an aerial reference map compared to prior art which similarly tests air-ground localization on the KITTI benchmark.
We report the 2D localization error for both stereo odometry and the ground truth odometry for each of the five tested sequences.
Using ground truth odometry provides the maximum achievable accuracy of our pipeline.
Overall, while prior art achieves good accuracy, these methods are restricted to urban environments.
Our pipeline was designed to work in both urban and non-urban environments and therefore makes no assumptions about roads or lane markings.
As a result, our pipeline leverages less information than competing approaches.

Our achieved accuracy is competitive to prior art in structured environments, as is our ability to localize in multiple trajectories.
We outperform \cite{floros2013openstreetslam} on Sequences 00 and 02, and \cite{yan2019global} on Sequences 00, 06, and 07.
While in Sequences 00, 02, and 09, our accuracy using stereo SLAM does not surpass \cite{brubaker2015map} or \cite{miller2021any}, these methods assume urban structure.
These error statistics demonstrate our comparable accuracy to other methods regardless of our strict and practical assumption of an unstructured environment.

In general, symmetry in reference maps is challenging for our pipeline.
If the geometry of objects in different regions of the reference map look similar, the pipeline may globally localize to the wrong pose.
It is the symmetry in the geometry of objects, not the symmetry in road structure, which affects the algorithm's success.
Sequence 06 exemplifies this idea, as the symmetry of the road structure makes this sequence highly challenging~\cite{brubaker2015map}.
However, despite this symmetry, our pipeline is able to successfully localize because there was little symmetry in parking space and traffic sign locations.

\section{Conclusions}\label{sec:conclusion}
We presented a pipeline for global localization and guided relocalization of a vehicle’s pose in unstructured environments using maps created from various veiwpoints.
Experiments with the Katwijk dataset and the KITTI benchmark demonstrate the pipeline's view-invariant property, robustness to outliers, and capability of localizing in unstructured environments.

\balance %

\bibliographystyle{IEEEtran}
\bibliography{refs}

\end{document}